\documentclass[10pt,twocolumn,letterpaper]{article}

\usepackage{cvpr}
\usepackage{times}
\usepackage{epsfig}
\usepackage{graphicx}
\usepackage{amsmath}
\usepackage{amssymb}
\usepackage{setspace}
\usepackage{bbm}

\newcommand{\bs}{\boldsymbol{s}}

\newcommand{\bmu}{\boldsymbol{\mu}}
\newcommand{\bsigma}{\boldsymbol{\sigma}}
\newcommand{\bphi}{\boldsymbol{\phi}}
\newcommand{\bpsi}{\boldsymbol{\psi}}
\newcommand{\bSigma}{\boldsymbol{\Sigma}}

\newcommand{\btheta}{\boldsymbol{\theta}}

\newcommand{\KL}{\text{KL}}

\newcommand{\bz}{\boldsymbol{z}}

\newcommand{\bx}{\boldsymbol{x}}

\usepackage{soul}

\newcommand{\Expect}{{\rm I\kern-.3em E}}

\usepackage[pagebackref=true,breaklinks=true,letterpaper=true,colorlinks,bookmarks=false]{hyperref}

\cvprfinalcopy 

\def\cvprPaperID{4335} 

\ifcvprfinal\fi
\begin{document}

\title{Anatomical Priors in Convolutional Networks for \\ Unsupervised Biomedical Segmentation}

\author{Adrian V. Dalca\\
MIT and MGH\\
{\tt\small adalca@mit.edu}
\and
John Guttag\\
MIT\\
{\tt\small guttag@mit.edu}
\and
Mert R. Sabuncu\\
Cornell University\\
{\tt\small msabuncu@cornell.edu}
}

\maketitle



\begin{abstract}
	
	We consider the problem of segmenting a biomedical image into anatomical regions of interest. We specifically address the frequent scenario where we have no paired training data that contains images and their manual segmentations. Instead, we employ unpaired segmentation images to build an anatomical prior. Critically these segmentations can be derived from imaging data from a different dataset and imaging modality than the current task. We introduce a generative probabilistic model  that employs the learned prior through a convolutional neural network to compute segmentations in an unsupervised setting. We conducted an empirical analysis of the proposed approach in the context of structural brain MRI segmentation, using a multi-study dataset of more than 14,000 scans.
	Our results show that an anatomical prior can enable fast unsupervised segmentation which is typically not possible using standard convolutional networks.
	The integration of anatomical priors can facilitate CNN-based anatomical segmentation in a range of novel clinical problems, where few or no annotations are available and thus standard networks are not trainable. 
	The code is freely available at \url{http://github.com/adalca/neuron}.

\end{abstract}

\section{Introduction}

Biomedical image segmentation plays a crucial role in many applications, such as population analysis, disease progression modelling, or treatment planning. Convolutional neural networks (CNNs), a class of deep learning methods, have been employed to derive powerful biomedical segmentation algorithms, showing promise of overcoming limitations in previous methods~\cite{badrinarayanan2015segnet,chen2016deep,ronneberger2015u,wachinger2017deepnat}. However, CNN-based approaches most often depend on (large-scale) training data, particularly in the form of image scans paired with segmentations. These annotations are often costly and challenging to obtain because they require the tedious effort of a trained expert, taking several expert hours per scan.

\subsection{Contributions}

To our knowledge, there has not been a theoretically rigorous effort to integrate rich probabilistic anatomical priors with a CNN-based segmentation model in a computationally effective manner. We introduce a generative model for biomedical segmentation that employs a deep anatomical prior. We describe a principled derivation that follows directly from our generative model. We demonstrate that this yields intuitive cost functions and simpler models. We use an auto-encoding variational CNN to characterize the anatomical prior, and an encoder-decoder CNN to provide fast segmentation of medical images in unsupervised settings. 

We demonstrate the method in an unsupervised biomedical image segmentation setting where paired annotations are not available. Our proposed strategy is general and computationally efficient, provides a natural framework for sampling possible subject-specific segmentations of a scan, and provides uncertainty estimates for these segmentations.



\section{Related Work}

\subsection{Segmentation Convolutional Neural Networks}

CNN-based segmentation approaches generally rely on fully convolutional architectures applied to image data. They extract hierarchical and multi-resolution features that are in turn combined to compute a semantic segmentation~\cite{milletari2016v,ronneberger2015u,roy2017error,wachinger2017deepnat}.

A popular discriminative segmentation architecture, U-net~\cite{ronneberger2015u}, involves a convolutional encoder or downsampling network, followed by a convolutional decoder or upsampling network, and skip-connections between layers. 
The encoder captures relevant features of the input image at different resolutions. 
The decoder then synthesizes a high-resolution segmentation, using the skip connections to achieve voxel-level precision. While the exact architecture of these networks, such as the number of layers and levels, size of convolution kernels, or application of batch normalization vary, they typically involve millions of parameters and necessitate large datasets and data augmentation techniques to train. 

CNN-based segmentation models have two major shortcomings: the dependency on annotated data, limiting their use in unsupervised settings; and their lack of anatomical knowledge. The latter limits the network's ability to be faithful to known anatomical shapes during segmentation.

In our work, we use CNN architectures to learn anatomical priors and segment medical images. The prior eliminates the burden of providing paired example segmentations.

\subsection{Priors for Convolutional Neural Networks}

A clinical expert performing manual delineation relies on spatial coordinates and prior knowledge about anatomy, and may use a template of the structures to constrain the task. This process draws on the anatomical similarity across patient scans. This is in stark contrast with typical computer vision problems that have led to many popular CNN architectures, where object location, shape, and appearance can be unpredictable. 

Convolutional methods are often limited in incorporating domain expertise. For example, U-Net~\cite{ronneberger2015u} and its derivatives produce segmentation algorithms that do not exploit location information or other explicit anatomical priors. A CNN might have difficulty differentiating two distinct objects that are consistently in two specific parts of the scan, if they have the same intensity and context (as in bilateral structures in two hemispheres)\footnote{Assuming that the field of view of the network is constrained to not include the other object's vicinity}. While increasingly more complex networks that extend receptive fields may tease out object differences in supervised settings, the problem would be trivial if we consider anatomical knowledge like spatial location.  Furthermore, in these modalities, image contrast can be weak or noisy in certain regions resulting in uncertainty of the segmentations. An anatomical prior can resolve these ambiguities, while making the segmentation task easier.

A popular strategy to explicitly employ prior structure in CNNs for biomedical image segmentation is to use a conditional random field (CRF) as a post-processing step~\cite{havaei2017brain,roth2015deeporgan,wachinger2017deepnat}. However, CRFs only capture local constraints, and add to the computational burden.  Location information has been included as a feature in patch-based CNN segmentation networks~\cite{wachinger2017deepnat}. While this addition carries prior location information, it is network-specific, increases the parameter burden on the network, and does not capture shape information.

Recent methods have employed shape priors for neural network solutions in supervised problems~\cite{oktay2017anatomically,ravishankar2017learning}. In particular, they often design a series of networks that learn representations of images and segmentations in a supervised setting. They propose \textit{ad-hoc} cost functions that encourage the computed segmentations to be similar to both the learned shape and the ground truth. These methods attempt to correct segmentations produced by standard CNNs by adding a prior constraint.

Convolutional image generative models, such as generative adversarial nets, have grown in popularity. They have recently been applied to biomedical image segmentations~\cite{isola2016image,moeskops2017adversarial} in a supervised setting where standard loss functions are combined with adversarial losses. A series of recent papers in the computer vision community removes the requirement for paired data by introducing a cycle dependency~\cite{zhu2017unpaired}. However, these methods are less applicable in medical image segmentation with many anatomical labels, as an image signal can pass through the rich networks at low cost, leading to a perfect cycle loss, circumventing the required constraints~\cite{zhu2017unpaired}.

Variational Bayes auto-encoders have been used for various tasks to learn probabilistic generative models, and often use convolutional networks~\cite{kingma2013}. Our method builds on these models to combine anatomical priors with image generation.




\subsection{Classical Generative Models}

Encoding and exploiting prior knowledge is common in generative models. Our inspiration comes from classical atlas-based probabilistic segmentation methods that estimate the maximum {\em{a posteriori}}~(MAP) probability based on a generative model involving a prior probability and likelihood ~\cite{colliot2006integration,fischl2002whole,kapur1996segmentation,patenaude2011bayesian,sabuncu2010generative,van1999automated,wells1996adaptive}.
%
%

The prior term captures knowledge of underlying anatomy and usually involves a probabilistic atlas and a spatial deformation that models geometric variation.
The spatial deformation can be explicitly solved using a registration algorithm or accounted for in a unified segmentation framework~\cite{ashburner2005unified}.

The likelihood models the physical process that yields medical image intensities, sometimes called the appearance model, conditioned on the latent anatomy.
These appearance models are often simpler, relying on additive and/or multiplicative Gaussian or Rician noise models~\cite{wells1996adaptive}.  
Model parameters are most often estimated using training data, such as annotated image pairs, for example using maximum likelihood. 

Given a new image, most popular segmentation algorithms use numerical non-convex optimization and can take several hours per image on a modern CPU.

In our model, we draw on ideas from classical model-based biomedical segmentation algorithms, convolutional neural networks (CNNs) used in semantic segmentation, and recent developments in variational Bayes approximations using neural networks. In our experiments, we consider the segmentation of structural brain MRI scans into cortical and subcortical regions of interest (ROIs). Our results show that the proposed anatomical prior enables rapid unsupervised segmentation. While complex, specialized tools exist for segmenting some specific scan modalities or particular diseases, they do not generalize to other modalities and can take hours to process one scan. Our goal is to provide a first general approach to biomedical image segmentation in an unsupervised setting.




\section{Generative Model}

\begin{figure}[t]
	\centering
	\begin{minipage}[t]{1\linewidth}
		\includegraphics[width=1\linewidth]{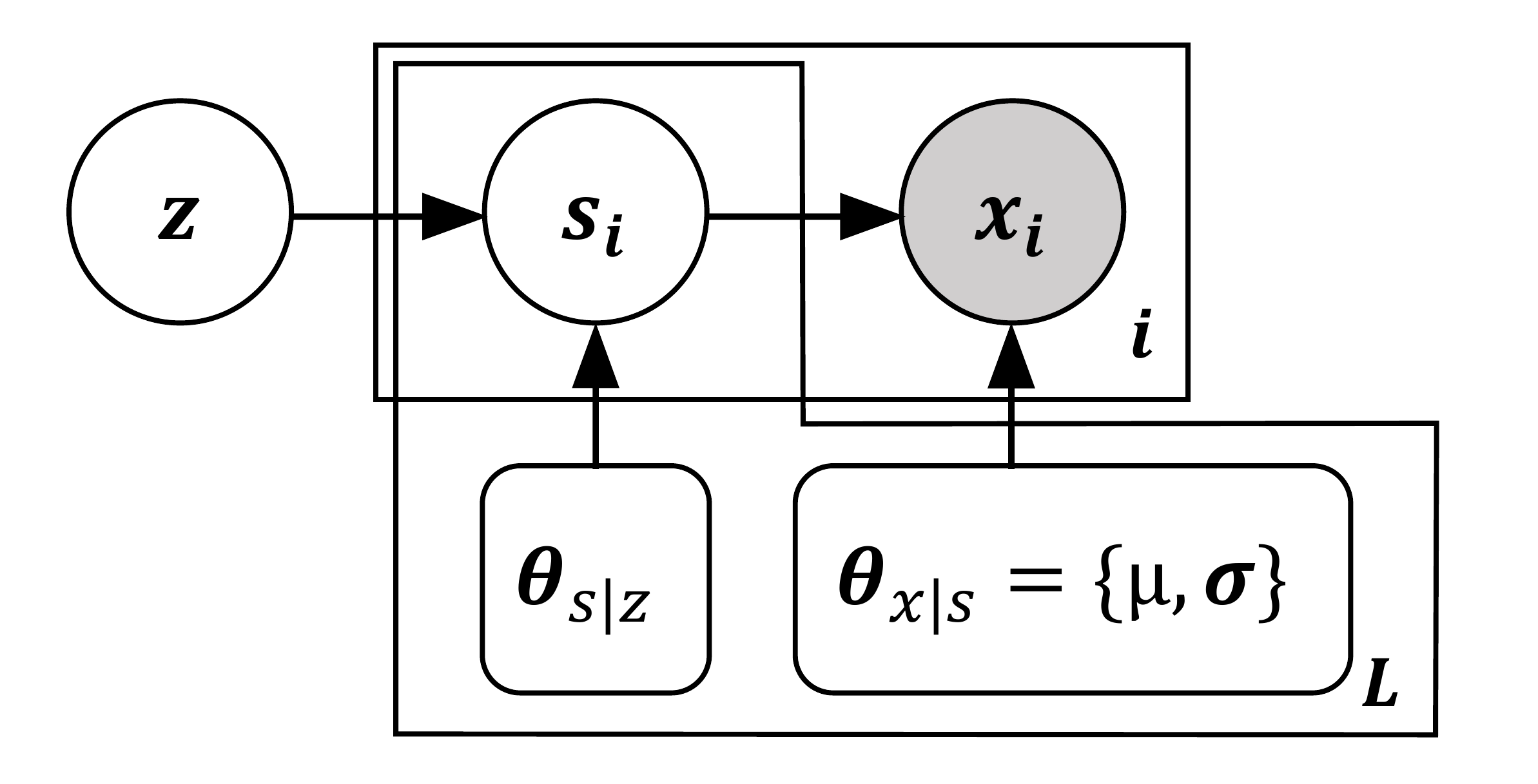}
	\end{minipage}
	\hfill
	%
	\begin{minipage}[b]{1\linewidth}
		\caption{A graphical representation of our generative model. Circles indicate random variables and rounded squares represent parameters. Shaded circles represent observed quantities and the plates indicate replication.~$\bx_i$ is the acquired image. The image intensities are generated from a normal distribution parametrized by~$\bmu_l$ and~$\sigma_l$ for each anatomical label~$l$ in the label map~$\bs$. Anatomical priors are controlled by the variable~$\bz$ and categorical parameters~$\btheta_{s|z}$.  }
		\label{fig:graphicalmodel_simple}
	\end{minipage}
\end{figure}

We let~$\bx$ be an (MR) 3D volume, and assume it is generated from a 3D anatomical segmentation map~$\bs$.  We will use~$x[j]$ and~$s[j]$ to denote the image intensity and label at voxel~$j$, respectively.

We use a generative model to describe the spatial distribution, shape, and appearance of anatomical structures. Figure~\ref{fig:graphicalmodel_simple} provides a graphical representation.

%
%
%



\begin{figure*}[t]
	\centering
	\makebox[\textwidth][c]{\includegraphics[width=1\textwidth]{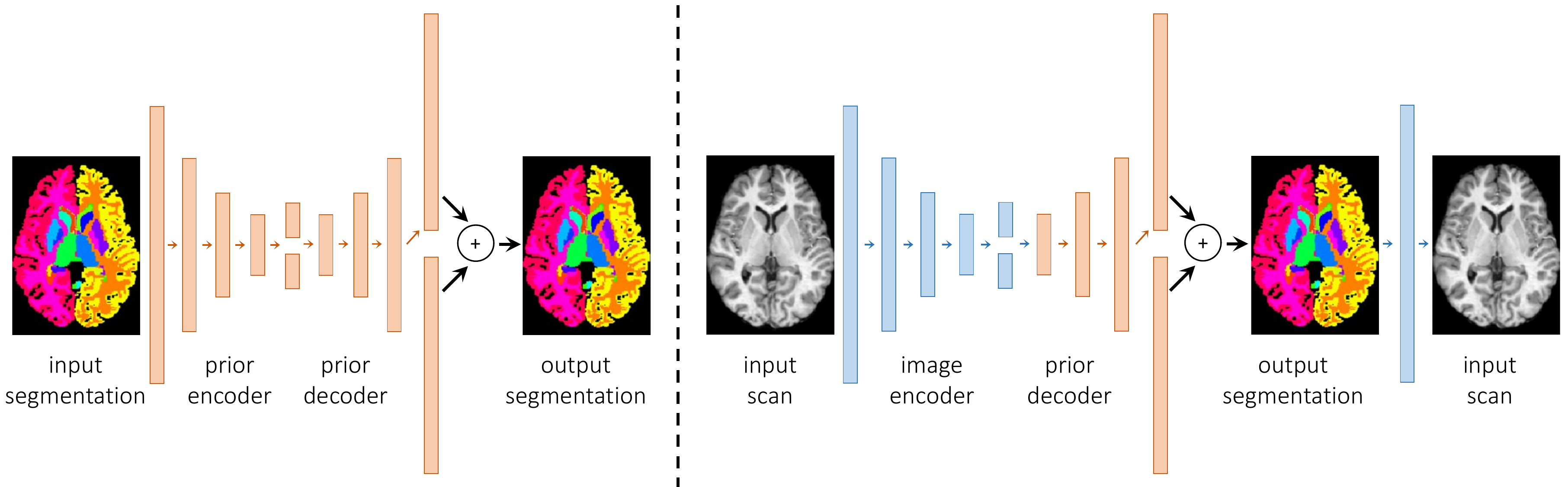}}
	\caption{Left: \textbf{Proposed Auto-Encoding Variational Anatomical Prior}. A variational auto-encoder is combined with a location-specific prior layer. Right: \textbf{Proposed architecture for learning generative model parameters}. Orange and blue arrows indicate down/up-sampling in the prior and full model, respectively, and rectangles represent a stack of convolutional layers with non-linearities, with their heights reflecting the size of the vectors.
	}
	\label{fig:UNet-Prior}
\end{figure*}

The prior captures our knowledge about spatial distributions and shape of anatomy. We let~$\bz$ be a latent variable representing an embedding of these shapes, and model the prior probability of this embedding as normal with mean~$\boldsymbol{0}$ and an identity covariance matrix:
\begin{align}
	p(\bz) = \mathcal{N}(\bz; \boldsymbol{0}, \boldsymbol{\mathbbm{1}}),
\end{align}
where $\mathcal{N}(\cdot;\bmu,\bSigma)$ is the normal distribution parametrized by mean~$\bmu$ and covariance~$\bSigma$.

We let~$\bs$ be drawn from a categorical prior distribution determined by the low-dimensional embedding~$\bz$ via~$p_{\btheta_{s|z}}(\bs|\bz)$:
\begin{align}
p_{\btheta_{s|z}}(\bs|\bz) &= \prod_j f_{j,s[j]}(\bz | \btheta_{s|z})
\label{eq:s_given_z}
\end{align}
where~$f_{j,l}(\cdot ; \btheta_{s|z})$ is the probability of label $l$ at voxel $j$.

Finally, given the label map~$\bs$, the intensity observations are generated via~$p_{\btheta_{x|s}}(\bx | \bs)$, sampled at each voxel from a normal distribution:
\begin{equation}
p_{\btheta_{x|s}}(\bx | \bs) = \prod_{j} \prod_{l} \mathcal{N}(\bx[j] ; \bmu_{l}, \sigma_{l}) ^ {\delta(\bs[j]=l)},
\label{eq:likelihood}
\end{equation}
where~$\btheta_{z|s}=\{\bmu_l, \sigma_l\}$, and~$\delta(\bs[j]=l)$ is the indicator function that evaluates to 1 if~$\bs[j]=l$ and 0 otherwise. The joint likelihood is therefore~\mbox{$p_{\btheta}(\bx, \bs | \bz) = p_{\btheta_{x|s}}(\bx | \bs) p_{\btheta_{s|z}}(\bs | \bz)$}, where~\mbox{$\btheta = \{\btheta_{s|z}, \btheta_{x|s}\}$}. Intuitively, the embedding~$\bz$ determines the possible anatomical shapes in~$\bs$, which in turn determine the possible observed images~$\bx$.

We describe the learning procedure in the next section. Given learned parameters, to obtain the segmentation~$\bs_i$ given a new image~$\bx_i$, we perform MAP estimation:
\begin{align}
\hat{\bs}_i &= \arg \max_{\bs_i} \log p(\bs_i|\bx_i; \btheta) \nonumber \\
&= \arg \max_{\bs_i} \log p(\bs_i, \bx_i; \btheta)
\label{eq:MAP1}
\end{align}

\section{Learning}

In this section, we describe a learning strategy that uses convolutional neural networks to estimate anatomical representations and optimize posterior segmentation distributions. This procedure is applicable to broad modelling choices for the probability distributions described above. 
We also discuss a separate learning procedure for the anatomical prior, uncertainty estimation, and implementation.

Without assuming voxel independence of the segmentation map given an image, estimating the posterior probability~$p_{\btheta}(\bs | \bx)$ is intractable since it involves integrating over the latent variable~$\bz$. Estimating~$p_{\btheta}(\bz|\bx,\bs)$ is similarly intractable, making the Expectation Maximization algorithm not pertinent. 

We first introduce an encoding probability~$q_{\bphi}(\bz|\bx, \bs)$ as an approximation to the intractable~$p_{\btheta}(\bz|\bx,\bs)$, similar to~\cite{kingma2013}. Consider the KL divergence between the approximate distribution~$q_{\bphi}(\bz|\bx, \bs)$ and the true posterior~$p_{\btheta}(\bz|\bx, \bs)$:
\begin{align}
\KL &\left[q_{\bphi}(\bz|\bx, \bs) || p_{\btheta}(\bz|\bx, \bs)  \right] \nonumber\\
  &= \Expect_{q} \left[ \log q_{\bphi}(\bz|\bx, \bs) - \log p_{\btheta}(\bz|\bx,\bs) \right] \\
  &= \Expect_{q} \left[ \log q_{\bphi}(\bz|\bx, \bs) - \log p_{\btheta}(\bx,\bs, \bz) \right] + \log p_{\btheta}(\bx, \bs).  \nonumber
\end{align}
Rearranging terms, we obtain
\begin{align}
	\log p(\bx, \bs) &= \KL \left[q_{\bphi}(\bz|\bx, \bs) || p_{\btheta}(\bz|\bx, \bs)  \right] \nonumber \\
	&+ \Expect_{q} \left[ \log p_{\btheta}(\bx,\bs, \bz) - \log q_{\bphi}(\bz|\bx, \bs) \right].
\end{align}
Since the KL divergence of the approximate and true posterior of~$\bz$ is non-negative, the second term is referred to as the \textit{variational lower bound} of the model evidence or joint probability. For a given approximate distribution~$q_{\bphi}(\bz|\bx, \bs)$, we can estimate $\btheta$ by optimizing the lower bound: 
%
\begin{align}
\mathcal{V}_{\text{model}}(&\btheta, \bphi; \bx, \bs) \nonumber \\
 &= \Expect_{q} \left[ \log p_{\btheta}(\bx,\bs, \bz) - \log q_{\bphi}(\bz|\bx, \bs) \right] \nonumber \\
 &= \Expect_{q} \left[ \log p_{\btheta}(\bx,\bs | \bz) \right] - \KL \left[  q_{\bphi}(\bz|\bx, \bs) ||  p(\bz)  \right]. 
 \label{eq:VLB}
\end{align}

We model the approximating posterior~$ q_{\bphi}(\bz|\bx, \bs)$ as a normal that depends on the image only:
\begin{align}
q_{\bphi}(\bz|\bx, \bs) &= q_{\bphi}(\bz|\bx) \nonumber \\
&= \mathcal{N}(\bz; \bmu_{z|x}, \bSigma_{z|x}).
\label{eq:posterior_z}
\end{align}
where~$\bSigma_{z|x}$ is diagonal.

We estimate the parameters of the approximating distribution using convolutional neural networks. We design an encoding convolutional neural network~$\text{enc}_{\bphi}(\bx)$ that takes as input~$\bx$ and outputs the parameters of the approximating posterior distribution~$\bmu_{z|x}(\bx)$, and~$\bSigma_{z|x}(\bx)$. This network learns how to embed an entire (MR) image into the most likely low-dimensional anatomical embedding~$\bz$ and its variance. 

Conditioned on~$\bz$, the probability of the segmentation can be computed with a \textit{decoder} network~$\text{dec}_{\btheta_{s|z}}(\bz)$ that takes~$\bz$ as input and outputs the parameters~$f(\bz;\btheta_{s|z})$ of the segmentation categorical distribution~$p_{\btheta_{s|z}}(\bs|\bz)$. The parameters~$\btheta_{s|z}$ of this decoder can be learned using a \textit{separate} set of segmentations, as described below. 

The final part of the generative model, the appearance or likelihood model, can also be learned with a neural network that takes a segmentation probability map as input and computes the parameters~$\bmu_l$. We separately estimate~$\sigma_l$, assuming additive zero mean Gaussian noise in an image, using a difference of Laplacian filters~\cite{immerkaer1996}. 


\subsection{Auto-Encoding Anatomical Prior}
In this work, we learn a prior independently from an unpaired segmentation dataset. This enables the flexibility of having an external description of the anatomy that need not be available in the current data. Unfortunately, as before, estimating the probability distribution~$p(\bs)$ is intractable. Following a derivation similar to the previous section and to the auto-encoding variational Bayes framework, we introduce an approximation~$q_{\bpsi}(\bz|\bs)$ to the posterior~$p(\bz|\bs)$ as a normal distribution:
\begin{align}
q_{\bpsi}(\bz|\bs) &= \mathcal{N}(\bz; \bmu_{z|s}, \bSigma_{z|s}),
\label{eq:posterior_z_s}
\end{align}
where~$\bSigma_{z|s}$ is diagonal, leading to the following lower bound:
\begin{align}
\mathcal{V}_\text{prior}(&\btheta, \bphi; \bs) \nonumber \\
&= \Expect_{q} \left[ \log p_{\btheta}(\bs, \bz) - \log q_{\bphi}(\bz|\bs) \right] \nonumber \\
&= \Expect_{q} \left[ \log p_{\btheta}(\bs | \bz) \right] - \KL \left[  q_{\bphi}(\bz|\bs) ||  p(\bz)  \right]. 
\end{align}
This can be optimized using a Stochastic Gradient Variational Bayes (SGVB) estimator that uses mini-batches. The reparametrization trick enables us to sample~$\bz_k \sim q_{\bphi}(\bz|\bs)$, leading to an approximation of the expectation~$\Expect\left[\cdot\right]$~\cite{kingma2013}. The loss~$\mathcal{L}_i(\btheta, \bphi ; \bs_i)$ for each data point~$\bs_i$ and sample~$\bz_k \sim q_{\bphi}(\bz|\bs_i)$ is
%
\begin{align}
\mathcal{L}_\text{prior}(&\btheta_{s|z}, \bpsi; \bs_i, \bz_k) \nonumber \\
&= \KL \left[ \log q_{\bpsi}(\bz | \bs_i) || \log p(\bz)  \right] - \log p_{\btheta_{s|z}}(\bs_i | \bz_k) \nonumber \\
&= \frac{1}{2} \sum_j \left(1 + \log \bSigma_{z|s_i}[j] - \bmu_{z|s_i}^2[j] - \bSigma_{z|s_i}[j] \right) \nonumber \\
&\quad- \sum_j \bs_i[j]\log f(\bz_k;\btheta_{s|z})[j].
\end{align}
We design an encoding network $\text{enc}_{\bpsi}(\bs)$ that takes a segmentation map as input and outputs the parameters~$\bmu_{z|s}$ and~$\bSigma_{z|s}$.
Importantly, we learn the parameters of the encoding network given only a set of segmentations~$\{\bs_i\}$, which can be derived from other imaging modalities and/or datasets.  
The segmentation prior therefore does not require paired training data in the traditional sense.
For example, we can use a prior computed using publicly available annotated datasets such as~\cite{klein2012101} in a problem that involves a different imaging modality than in the current task.

\subsection{Unsupervised Learning}


We assume we have learned a segmentation prior using the Auto-Encoding Anatomical Prior described in the previous section.  In particular, we will utilize the \textit{decoder} component of the prior model, namely~$p_{\btheta_{s|z}}(\bs|\bz)$.




If we had annotated pairs~$\{\bx_i,\bs_i\}$, we could jointly learn model parameters~$\btheta_{x|s}$, and variational parameters~$\bphi$ by optimizing the evidence lower bound objective~\eqref{eq:VLB}, similar to the previous section. For each sample~$\{\bx_i, \bs_i\}$ and sample~$\bz_k \sim q_{\bphi}(\bz|\bx_i,\bs_i)$, the loss function would be
\begin{align}
\mathcal{L}&_\text{model}(\btheta, \bphi ; \bx_i, \bs_i, \bz_k) = - \mathcal{V}_i(\btheta, \bphi ; \bx_i, \bs_i, \bz_k) \nonumber\\
&= \KL \left[ q_{\bphi}(\bz|\bx_i) || p(\bz)  \right] - \log p_{\btheta_{s|z}}(\bx_i, \bs_i | \bz_k). \nonumber \\
&= \KL \left[ q_{\bphi}(\bz|\bx_i) || p(\bz)  \right] - \log p_{\btheta_{s|z}}(\bs_i | \bz_k) \nonumber \\ &\quad - \log p_{\btheta_{x|s}}(\bx_i | \bs_i). \nonumber \\
&= \frac{1}{2} \sum_j \left(1 + \log \bSigma_{z|x_i}[j] - \bmu_{z|x_i}^2[j] - \bSigma_{z|x_i}[j] \right) \nonumber \\
&\quad  - \sum_j \bs_i[j]\log f(\bz;\btheta_{s|z})[j]  \label{eq:seg-loss-expanded} \\
&\quad  + \sum_j \sum_l \frac{\delta(\bs_i[j]=l)}{2 \sigma_l^2}  (\bx_i - \bmu_l), \nonumber
\end{align}
resulting in terms of KL divergence, segmentation map categorical cross-entropy, and intensity-based mean squared error, respectively. During training, these terms would ensure that the probability~$q_{\bphi}(\bz|\bx_i)$ stays close to the standard normal, while explaining the segmentations, and that the model parameters $\btheta_{x|s} = \{\bmu_l, \bsigma_l\} $ capture the relationship between the segmentations and the images. 

However, in this paper we tackle the unsupervised setting, where annotated pairs~$\{\bx_i,\bs_i\}$ are not available, and we only have the images~$\{\bx_i\}$. Therefore, we cannot compute the categorical cross entropy term in~\eqref{eq:seg-loss-expanded}. 
%
Instead, we marginalize over the segmentation~$\bs$ in the second term of the variational lower bound~\eqref{eq:VLB}:
\begin{align}
\Expect_{q}  & \left[ \log \int_{\bs} p_{\btheta}(\bx,\bs | \bz) d\bs \right] \nonumber \\
&= \Expect_{q} \left[ \log \int_{\bs} p_{\btheta}(\bx |\bs) p_{\btheta_{s|z}}(\bs | \bz) d\bs \right] \nonumber \\
&\ge  \Expect_{q} \left[ \int_{\bs} p_{\btheta_{s|z}}(\bs | \bz) \log p_{\btheta}(\bx |\bs)  d\bs \right] \nonumber \\
&=  \Expect_{q, p(\bs|\bz)} \left[  \log p_{\btheta}(\bx |\bs) \right] 
\end{align}
where we used Jensen's inequality. We therefore arrive at the following upper bound of the loss function:
%
%
\begin{align}
\mathcal{L}_\text{model}(&\btheta_{x|s}, \bphi ; \bx_i, \bz_k) \nonumber\\
&= \frac{1}{2} \sum_j \left(1 + \log \bSigma_{z|x_i}[j] - \bmu_{z|x_i}^2[j] - \bSigma_{z|x_i}[j] \right) \nonumber \\
&\quad  + \sum_j \sum_l \frac{f_{j,l}(\bz_k|\theta_{s|z})}{2 \sigma_l^2}  (\bx_i - \bmu_l),
\label{eq:seg-loss-expanded-nocat}
\end{align}
where we used the factorization of~$p_{\theta{s|z}}(\bs|\bz)$ over voxels from \eqref{eq:s_given_z}, and sample~$\bz_k \sim q_{\bphi}(\bz|\bx_i)$.


\subsection{Inference and uncertainty}

Given a new image~$\bx$, we approximate~\mbox{$\widehat{\bs} = \arg\max_{\bs} p_{\btheta}(\bs|\bx)$} by first obtaining~$\bmu_z$ using the encoder~$\text{enc}_{\btheta_{z|x}}(\bx)$, and taking the maximum segmentation at each voxel~$\widehat{\bs} = \arg\max_{\bs} \text{dec}_{\btheta_{s|z}}(\bmu_z)$. The operations are fast, since both are feed-forward neural networks.

This model also enables sampling segmentations conditioned on a particular image and enables estimation of uncertainty.  Given an input image~$\bx_i$, we can create samples~$\bz_k \sim q_{\bphi}(\bz | \bx)$ and~$\bs_k \sim p_{\btheta_{s|z}}(\bs| \bz_k)$, simulating different plausible segmentations for a given subject. 
We can estimate the uncertainty of our segmentation given a new image~$\bx_i$ using
\begin{align}
	H(\bs[j]) &= \Expect[-\log(p(\bs[j]|\bx_i))]  \\
	&= -\sum_l p(\delta(\bs[j]=l)|\bx_i) \log(p(\delta(\bs[j]=l)|\bx_i)). \nonumber
\end{align}
%

%
%
%
%
%

\begin{figure}[t]
	\centering
	\includegraphics[width=1\linewidth]{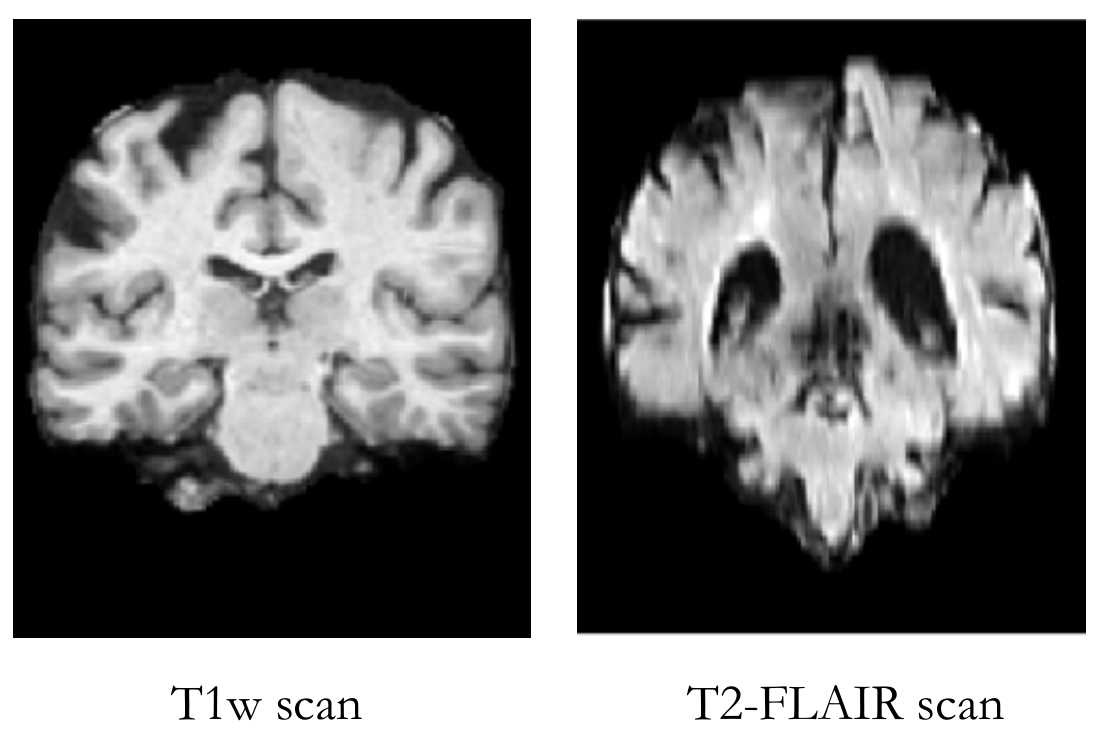}
	\caption{\textbf{Example T1w and T2-FLAIR images} highlighting the difference in anatomical differences, tissue contrast and scan quality.}
	\label{fig:example_slices}
\end{figure}

\subsection{Implementation}


A CNN can be seen as a hierarchical function, a set of concatenated functions, or layers. For example, CNNs often map some input image $\bx$ to an output probability $\hat{\bs}_p$:
\begin{equation}
\hat{\bs}_p = \boldsymbol{f}_{L} \circ \ldots \circ \boldsymbol{f}_{1}(\bx), \label{eq:CNN}
\end{equation}
where $\circ$ denotes concatenation, $\boldsymbol{f}_{i}$ is often some nonlinear function such as a rectified linear unit or ReLU or max-pool~\cite{goodfellow2016deep} applied to (linear) convolutions of the output of the previous layer $\boldsymbol{f}_{i-1}$ (with $\boldsymbol{f}_{0} = \bx$). 

Although we operate on 3D images, we use a 2D architecture in our experiments. We experimented with 3D architectures as well, but found little gain while facing significant challenges related to limitations between GPU memory, batch size, the number of features, and the number of labels possible. Each encoder consists of five downsampling levels of one convolution layer each, with 3x3 convolution kernels with elu activations, and 32 features for each kernel. The final layer is dense, with~$1000$-long encoding of the means and standard deviations representations. 

The decoder is a mirror of this design, but upsamples instead of downsampling and ends with a sigmoid activation. In addition, we use a final layer that implements a pixel-wise spatially-varying voxel-wise (location) prior $p_{\text{loc}}(\bs)$, which is multiplied with dec$(\bz)$ (in practice, we add the logarithms). 
As is common in the atlas-based segmentation literature, the prior $p_{\text{loc}}(\bs)$ was computed as the frequency of labels in the held out prior dataset, in affine-normalized coordinate system. This layer discourages any extreme decodings of~$\bz$ but does not capture shape properties, which is encoded in dec$(\bz)$.

We implement the normal probability~$p(\bx|\bs)$ with a single linear layer. We also find it useful to pre-train the image encoder using an image variational auto-encoder similar to the segmentation one. The encoder weights are used as initialization only. During training, we used the Adadelta optimizer~\cite{zeiler2012adadelta}. 

For the latent encoding layers representing~$\bmu_z$ and~$\bSigma_z$, we introduce an activation function  that discourages the sample activations from being too large, helping limit numerical issues stemming in sampling from these layers during the reparametrization trick. We use concepts from the softsign and tanh activations to define our function as~$\text{act}(\bx) = \text{softsign}_{\alpha}(\bx) \log (2 + \alpha * |\bx|)$. 


\section{Experiments}

We demonstrate our model on two datasets. For the first dataset, we obtain ground truth segmentations using a specialized algorithm with intense computational requirements, combined with manual work and QC~\cite{fischl2012}. We use a subset of these segmentations to learn the prior probability parameters. We treat the rest of the dataset as \textit{unsupervised}, where we only use the ground truth segmentations as validation. For the second dataset we do not have ground truth, offering a realistic scenario. Figure~\ref{fig:example_slices} shows example images from the two datasets, highlighting the difference, and the difficulty of the task.

\subsection{Data}

\begin{figure*}[t]
	\centering
	\includegraphics[width=1\linewidth]{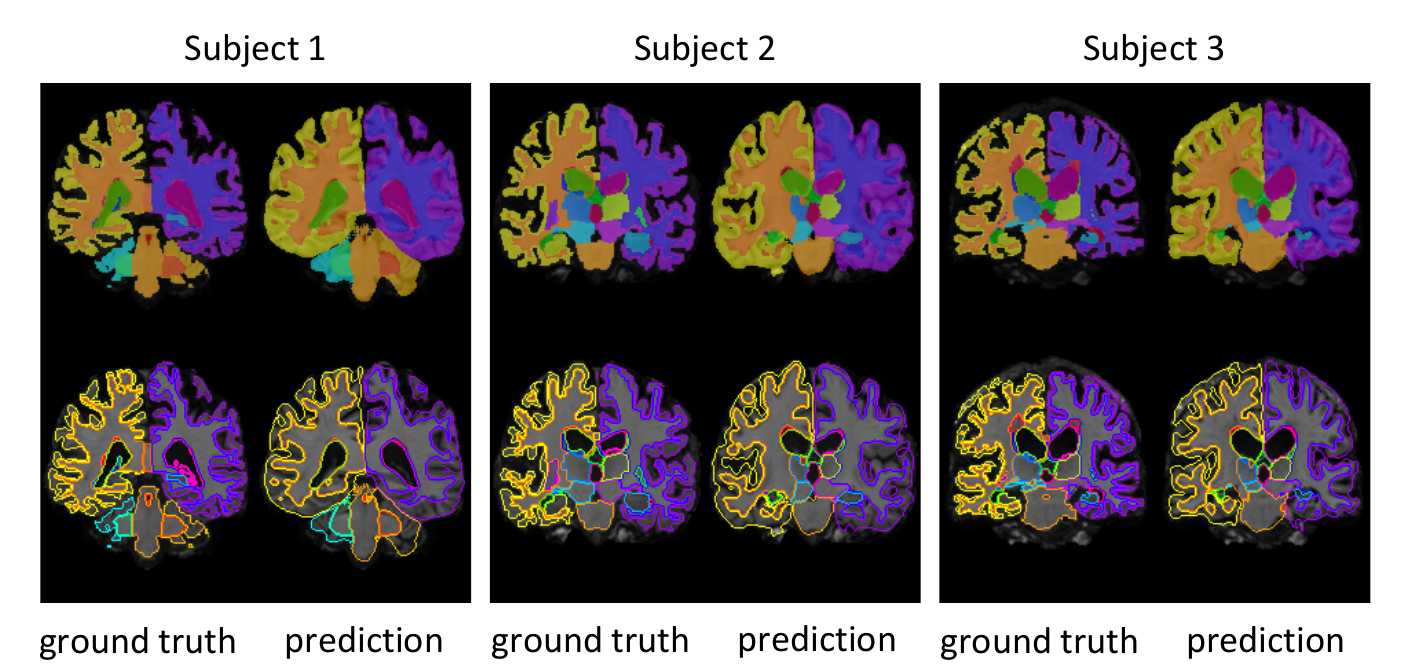}
	\caption[Segmentation Results]{\textbf{T1 segmentation examples}. For each subject, the left column shows the "ground truth" as estimated with FreeSurfer, and the right illustrates our prediction. The first row overlays anatomical  structures  on top of the subject scan to clearly indicate the proposed segmentation. The second row shows outlines of each structure to allow comparison with the subject scan. }
	\label{fig:example-T1w}
\end{figure*}

\begin{figure}[t]
	\centering
	\includegraphics[width=1\linewidth]{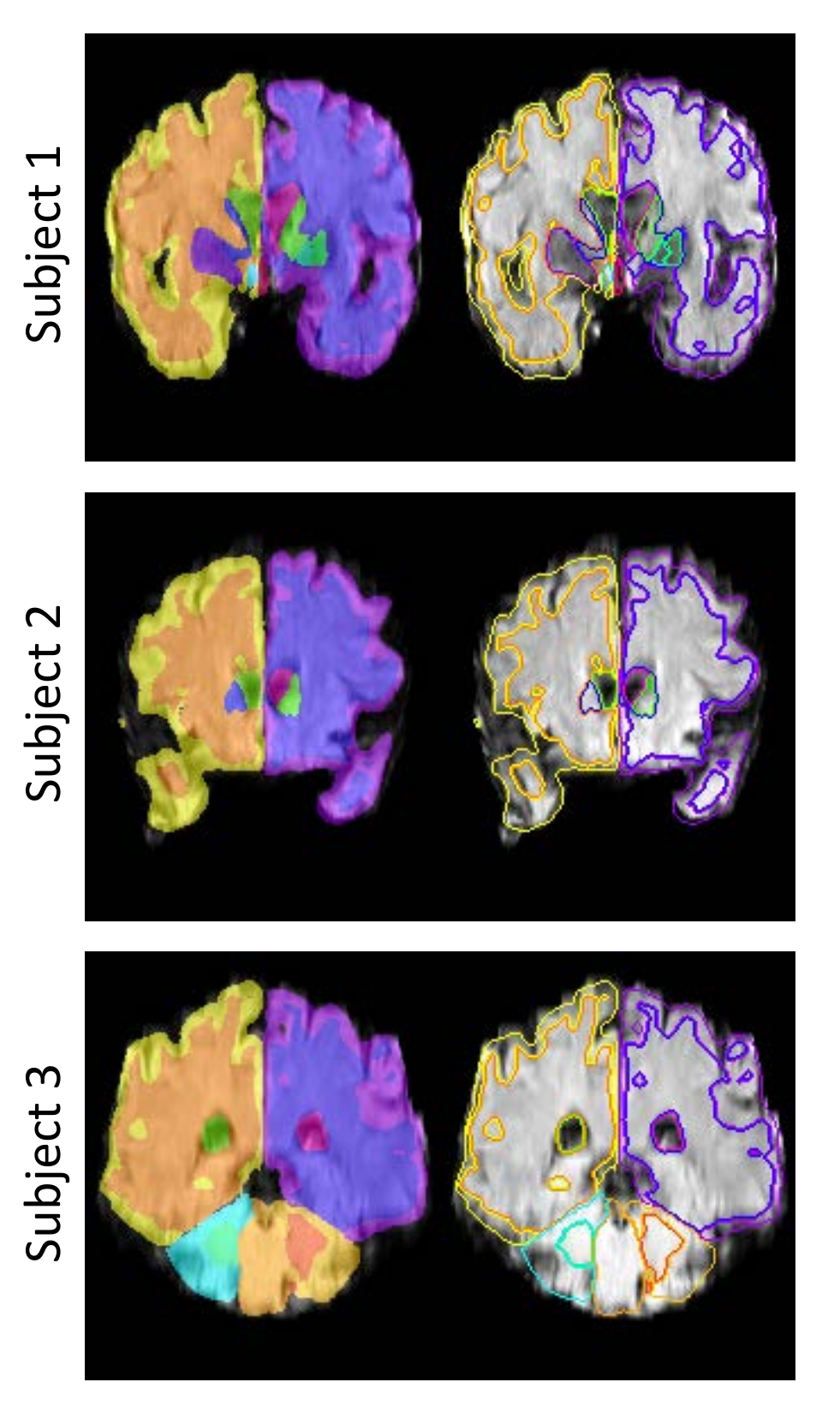}
	\caption[Segmentation Results]{\textbf{T2-FLAIR segmentation examples}. The first column overlays anatomical  structures  on top of the subject scan to clearly indicate the proposed segmentation. The second column draws outlines of each structure to allow comparison with the subject scan. Different coronal slices illustrate the variability and difficulty of the task.}
	\label{fig:example-T2}
\end{figure}

\subsubsection*{T1w scan dataset}

We gathered a large-scale multi-site, multi-study dataset of more than 14,000 T1-weighted brain MRI scans from eight publicly available datasets:  including data from ADNI~\cite{mueller2005ways}, OASIS~\cite{marcus2007open}, ABIDE~\cite{di2014autism}, ADHD200~\cite{milham2012adhd}, MCIC~\cite{gollub2013mcic}, PPMI~\cite{marek2011parkinson}, HABS~\cite{dagley2015harvard}, and Harvard GSP~\cite{holmes2015brain}. Subject age ranges, health states, and acquisition details vary with each dataset, but all scans were resampled to a 256x256x256 grid with 1mm isotropic voxels, and all images cropped to 160x192x224 to eliminate entirely-background voxels.

We carry out standard pre-processing steps, including affine spatial normalization using FreeSurfer for each scan~\cite{fischl2012}. All MRIs were also segmented with FreeSurfer - a task that takes several CPU hours per scan. We also applied quality control (QC) using visual inspection to catch gross errors in segmentation results. 

We partitioned the data into a prior training subset of 5,000 images, where we only used the annotations. The rest of the data was treated as an unannotated dataset, where QCed segmentations were only used for validation. 

While developing the network architectures we partitioned the rest of the data into training, validation and test sets. Once the architecture was fixed, we reported results on the test dataset by training and evaluating the model in an \textit{unsupervised} fashion.

\subsubsection*{T2-FLAIR scan dataset}

We also gathered a dataset of more than 3800 T2-FLAIR scans, a significantly different MR modality, from the ADNI cohort.  These scans exhibit significantly different tissue properties compared to the T1w images, lower acquisition quality, and exhibit~$5$mm slice spacing (Figure~\ref{fig:example_slices}). They provide a good test of our hypothesis that priors learned are useful for segmenting image data with different tissue properties. To our knowledge there is no automatic method to obtain detailed anatomical segmentations for these images. We affinely align these images to the same space as the T1 images using mutual information based affine registration with ANTs~\cite{avants2009advanced}. We perform brain extraction using an in-house developed neural network-based algorithm that uses a UNet architecture and extensive data augmentation. 

In the set of annotations that we used to train the prior, we avoided including any annotations coming from ADNI subjects whose T2-FLAIR scans are in this dataset.

\begin{figure}[t]
	\centering
	\includegraphics[width=1\linewidth]{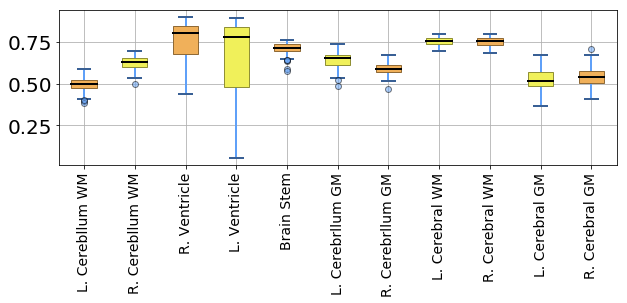}
	\caption[Dice Graph]{\textbf{Volume Overlap} measured via (Dice) for several structures in the T1w images.}
	\label{fig:dice}
\end{figure}

\subsection{Evaluation}

We evaluate our results both visually and quantitatively. For the \textit{T1w dataset}, we use a volume overlap measure, Dice, to quantify the automatic segmentation results~\cite{dice1945measures}:
\begin{align}
\text{Dice}(\widehat{\bs}, \bs_t) = \frac{2 \sum_j \delta(\widehat{\bs}[j]=l) \delta({\bs}_t[j]=l)}{\sum_j \delta(\widehat{\bs}[j]=l) + \sum_j\delta({\bs}_t[j]=l)}.
\end{align}
where~$\widehat{\bs}$ is the predicted segmentation map, and~$\bs_t$ indicates the ground truth (FreeSurfer) label at each location. A Dice score of 1 indicates perfect segmentation.

We experimented segmenting in the unsupervised setting with standard UNet architectures, using the image MSE and mutual information loss functions. Because of the many structures that share similar intensities, these architectures are not able to produce sensible segmentations that resemble the correct segmentations, and we omit them from these results. 
Classical unsupervised methods that include sophisticated prior anatomical information take a significant amount time to run, and for T1w we regard FreeSurfer results as an optimistic bound for the T1w data. However, as these methods tend to be focused on specific modalities, there is no annotation tool for  cortical and subcortical regions in T2-FLAIR. We evaluate the T2-FLAIR segmentation visually in Figure~\ref{fig:example-T2}. 




%


\subsection{Results}

At test time, a new subject only needs to be affinely registered to a template, after which the proposed CNN model evaluates a segmentation estimate. The entire process takes less than a few seconds on an NVidia Titan X GPU. 

Fig.~\ref{fig:example-T1w} shows a series of example segmentations for the T1w dataset demonstrating that our method is able to estimate approximate anatomical structures, reproducing the general location as well as the shape of structures. Fine details, such as details of cortex folding, is not easily captured by the prior encoding, leading to smooth segmentation predictions. Fig.~\ref{fig:dice} illustrates the average Dice measure across several anatomical regions for T1 scans. We focus on the most prevalent (larger) structures, which can also be evaluated in detail in the visualizations of Figure~\ref{fig:example-T1w}. 

Fig.~\ref{fig:example-T2} shows results of our algorithm on T2-FLAIR scans. Even with the significantly lower image quality and different tissue contrasts, our algorithm 
is able to use the prior information to predict plausible segmentations, even given challenging images in this unsupervised scenario.

\subsection{Discussion}

Our method is able to detect anatomical structures that are guided by image contrast while respecting anatomical shapes according to the prior.  Rapid, zero-shot segmentation is a  challenging task, and has not been tackled by previous methods.  As such, the absence of prior results makes it difficult to fully interpret current results. The detailed FreeSurfer results are an upper bound, which any model is unlikely to achieve in the unsupervised setting. We omit showing results from lower bound (simplistic) baselines, such as the unsupervised U-Net model described above, since these models yielded nonsensical segmentations. To the best of our knowledge, our results are the first for zero-shot neural-network based segmentation of brain structures.

\section{Conclusion}
In this paper, we introduced a generative probabilistic model  that employs a prior model learned through a convolutional neural network to compute segmentations in an unsupervised setting. We can interpret the anatomical prior as encouraging the neural network to predicting segmentation maps that come from a known distribution characterized by~$\bz$ while simultaneously producing images that agree with the observed scan.
We demonstrate that our model enables segmentation using convolutional networks leading to rapid inference in a setting where segmentation is traditionally not possible, or takes hours to obtain for a single scan. The integration of priors promises to facilitate accurate anatomical segmentation in a variety of novel clinical problems with limited dataset availability.

{\small
\bibliographystyle{ieee}
\bibliography{bibliography}

\begin{thebibliography}{10}\itemsep=-1pt

\bibitem{ashburner2005unified}
J.~Ashburner and K.~J. Friston.
\newblock Unified segmentation.
\newblock {\em Neuroimage}, 26(3):839--851, 2005.

\bibitem{avants2009advanced}
B.~B. Avants, N.~Tustison, and G.~Song.
\newblock Advanced normalization tools (ants).
\newblock {\em Insight j}, 2:1--35, 2009.

\bibitem{badrinarayanan2015segnet}
V.~Badrinarayanan, A.~Kendall, and R.~Cipolla.
\newblock Segnet: A deep convolutional encoder-decoder architecture for image
  segmentation.
\newblock {\em arXiv preprint arXiv:1511.00561}, 2015.

\bibitem{chen2016deep}
H.~Chen, X.~J. Qi, J.~Z. Cheng, and P.~A. Heng.
\newblock Deep contextual networks for neuronal structure segmentation.
\newblock In {\em Thirtieth AAAI conference on artificial intelligence}, 2016.

\bibitem{colliot2006integration}
O.~Colliot, O.~Camara, and I.~Bloch.
\newblock Integration of fuzzy spatial relations in deformable
  models—application to brain {MRI} segmentation.
\newblock {\em Pattern recognition}, 39(8):1401--1414, 2006.

\bibitem{dagley2015harvard}
A.~Dagley, M.~LaPoint, W.~Huijbers, T.~Hedden, D.~G. McLaren, J.~P. Chatwal,
  K.~V. Papp, R.~E. Amariglio, D.~Blacker, D.~M. Rentz, et~al.
\newblock Harvard aging brain study: dataset and accessibility.
\newblock {\em NeuroImage}, 2015.

\bibitem{di2014autism}
A.~Di~Martino, C.-G. Yan, Q.~Li, E.~Denio, F.~X. Castellanos, K.~Alaerts, J.~S.
  Anderson, M.~Assaf, S.~Y. Bookheimer, M.~Dapretto, et~al.
\newblock The autism brain imaging data exchange: towards a large-scale
  evaluation of the intrinsic brain architecture in autism.
\newblock {\em Molecular psychiatry}, 19(6):659--667, 2014.

\bibitem{dice1945measures}
L.~R. Dice.
\newblock Measures of the amount of ecologic association between species.
\newblock {\em Ecology}, 26(3):297--302, 1945.

\bibitem{fischl2012}
B.~Fischl.
\newblock Freesurfer.
\newblock {\em Neuroimage}, 62(2):774--781, 2012.

\bibitem{fischl2002whole}
B.~Fischl, D.~H. Salat, E.~Busa, M.~Albert, M.~Dieterich, C.~Haselgrove, A.~Van
  Der~Kouwe, R.~Killiany, D.~Kennedy, S.~Klaveness, et~al.
\newblock Whole brain segmentation: automated labeling of neuroanatomical
  structures in the human brain.
\newblock {\em Neuron}, 33(3):341--355, 2002.

\bibitem{gollub2013mcic}
R.~L. Gollub, J.~M. Shoemaker, M.~D. King, T.~White, S.~Ehrlich, S.~R.
  Sponheim, V.~P. Clark, J.~A. Turner, B.~A. Mueller, V.~Magnotta, et~al.
\newblock The mcic collection: a shared repository of multi-modal, multi-site
  brain image data from a clinical investigation of schizophrenia.
\newblock {\em Neuroinformatics}, 11(3):367--388, 2013.

\bibitem{goodfellow2016deep}
I.~Goodfellow, Y.~Bengio, and A.~Courville.
\newblock {\em Deep learning}.
\newblock MIT Press, 2016.

\bibitem{havaei2017brain}
M.~Havaei, A.~Davy, D.~Warde-Farley, A.~Biard, A.~Courville, Y.~Bengio, C.~Pal,
  P.-M. Jodoin, and H.~Larochelle.
\newblock Brain tumor segmentation with deep neural networks.
\newblock {\em Medical image analysis}, 35:18--31, 2017.

\bibitem{holmes2015brain}
A.~J. Holmes, M.~O. Hollinshead, T.~M. O’Keefe, V.~I. Petrov, G.~R. Fariello,
  L.~L. Wald, B.~Fischl, B.~R. Rosen, R.~W. Mair, J.~L. Roffman, et~al.
\newblock Brain genomics superstruct project initial data release with
  structural, functional, and behavioral measures.
\newblock {\em Scientific data}, 2, 2015.

\bibitem{immerkaer1996}
J.~Immerkaer.
\newblock Fast noise variance estimation.
\newblock {\em Computer vision and image understanding}, 64(2):300--302, 1996.

\bibitem{isola2016image}
P.~Isola, J.-Y. Zhu, T.~Zhou, and A.~A. Efros.
\newblock Image-to-image translation with conditional adversarial networks.
\newblock {\em arXiv preprint arXiv:1611.07004}, 2016.

\bibitem{kapur1996segmentation}
T.~Kapur, W.~E.~L. Grimson, W.~M. Wells, and R.~Kikinis.
\newblock Segmentation of brain tissue from magnetic resonance images.
\newblock {\em Medical image analysis}, 1(2):109--127, 1996.

\bibitem{kingma2013}
D.~P. Kingma and M.~Welling.
\newblock Auto-encoding variational bayes.
\newblock {\em arXiv preprint arXiv:1312.6114}, 2013.

\bibitem{klein2012101}
A.~Klein and J.~Tourville.
\newblock 101 labeled brain images and a consistent human cortical labeling
  protocol.
\newblock {\em Frontiers in neuroscience}, 6:171, 2012.

\bibitem{marcus2007open}
D.~S. Marcus, T.~H. Wang, J.~Parker, J.~G. Csernansky, J.~C. Morris, and R.~L.
  Buckner.
\newblock Open access series of imaging studies (oasis): cross-sectional mri
  data in young, middle aged, nondemented, and demented older adults.
\newblock {\em Journal of cognitive neuroscience}, 19(9):1498--1507, 2007.

\bibitem{marek2011parkinson}
K.~Marek, D.~Jennings, S.~Lasch, A.~Siderowf, C.~Tanner, T.~Simuni, C.~Coffey,
  K.~Kieburtz, E.~Flagg, S.~Chowdhury, et~al.
\newblock The parkinson progression marker initiative (ppmi).
\newblock {\em Progress in neurobiology}, 95(4):629--635, 2011.

\bibitem{milham2012adhd}
M.~P. Milham, D.~Fair, M.~Mennes, S.~H. Mostofsky, et~al.
\newblock The adhd-200 consortium: a model to advance the translational
  potential of neuroimaging in clinical neuroscience.
\newblock {\em Frontiers in systems neuroscience}, 6:62, 2012.

\bibitem{milletari2016v}
F.~Milletari, N.~Navab, and S.-A. Ahmadi.
\newblock V-net: Fully convolutional neural networks for volumetric medical
  image segmentation.
\newblock In {\em 3D Vision (3DV), 2016 Fourth International Conference on},
  pages 565--571. IEEE, 2016.

\bibitem{moeskops2017adversarial}
P.~Moeskops, M.~Veta, M.~W. Lafarge, K.~A. Eppenhof, and J.~P. Pluim.
\newblock Adversarial training and dilated convolutions for brain mri
  segmentation.
\newblock In {\em Deep Learning in Medical Image Analysis and Multimodal
  Learning for Clinical Decision Support}, pages 56--64. Springer, 2017.

\bibitem{mueller2005ways}
S.~G. Mueller, M.~W. Weiner, L.~J. Thal, R.~C. Petersen, C.~R. Jack, W.~Jagust,
  J.~Q. Trojanowski, A.~W. Toga, and L.~Beckett.
\newblock Ways toward an early diagnosis in {Alzheimer’s disease: the
  Alzheimer’s Disease Neuroimaging Initiative (ADNI)}.
\newblock {\em Alzheimer's \& Dementia}, 1(1):55--66, 2005.

\bibitem{oktay2017anatomically}
O.~Oktay, E.~Ferrante, K.~Kamnitsas, M.~Heinrich, W.~Bai, J.~Caballero,
  R.~Guerrero, S.~Cook, A.~de~Marvao, D.~O'Regan, et~al.
\newblock Anatomically constrained neural networks (acnn): Application to
  cardiac image enhancement and segmentation.
\newblock {\em arXiv preprint arXiv:1705.08302}, 2017.

\bibitem{patenaude2011bayesian}
B.~Patenaude, S.~M. Smith, D.~N. Kennedy, and M.~Jenkinson.
\newblock A {Bayesian} model of shape and appearance for subcortical brain
  segmentation.
\newblock {\em Neuroimage}, 56(3):907--922, 2011.

\bibitem{ravishankar2017learning}
H.~Ravishankar, R.~Venkataramani, S.~Thiruvenkadam, P.~Sudhakar, and V.~Vaidya.
\newblock Learning and incorporating shape models for semantic segmentation.
\newblock In {\em International Conference on Medical Image Computing and
  Computer-Assisted Intervention}, pages 203--211. Springer, 2017.

\bibitem{ronneberger2015u}
O.~Ronneberger, P.~Fischer, and T.~Brox.
\newblock U-net: Convolutional networks for biomedical image segmentation.
\newblock In {\em International Conference on Medical Image Computing and
  Computer-Assisted Intervention}, pages 234--241. Springer, 2015.

\bibitem{roth2015deeporgan}
H.~R. Roth, L.~Lu, A.~Farag, H.-C. Shin, J.~Liu, E.~B. Turkbey, and R.~M.
  Summers.
\newblock Deeporgan: Multi-level deep convolutional networks for automated
  pancreas segmentation.
\newblock In {\em International Conference on Medical Image Computing and
  Computer-Assisted Intervention}, pages 556--564. Springer, 2015.

\bibitem{roy2017error}
A.~G. Roy, S.~Conjeti, D.~Sheet, A.~Katouzian, N.~Navab, and C.~Wachinger.
\newblock Error corrective boosting for learning fully convolutional networks
  with limited data.
\newblock {\em arXiv preprint arXiv:1705.00938}, 2017.

\bibitem{sabuncu2010generative}
M.~R. Sabuncu, B.~T. Yeo, K.~Van~Leemput, B.~Fischl, and P.~Golland.
\newblock A generative model for image segmentation based on label fusion.
\newblock {\em IEEE transactions on medical imaging}, 29(10):1714--1729, 2010.

\bibitem{van1999automated}
K.~Van~Leemput, F.~Maes, D.~Vandermeulen, and P.~Suetens.
\newblock Automated model-based tissue classification of {MR} images of the
  brain.
\newblock {\em IEEE transactions on medical imaging}, 18(10):897--908, 1999.

\bibitem{wachinger2017deepnat}
C.~Wachinger, M.~Reuter, and T.~Klein.
\newblock Deepnat: Deep convolutional neural network for segmenting
  neuroanatomy.
\newblock {\em NeuroImage}, 2017.

\bibitem{wells1996adaptive}
W.~M. Wells, W.~E.~L. Grimson, R.~Kikinis, and F.~A. Jolesz.
\newblock Adaptive segmentation of {MRI} data.
\newblock {\em IEEE transactions on medical imaging}, 15(4):429--442, 1996.

\bibitem{zeiler2012adadelta}
M.~D. Zeiler.
\newblock Adadelta: an adaptive learning rate method.
\newblock {\em arXiv preprint arXiv:1212.5701}, 2012.

\bibitem{zhu2017unpaired}
J.-Y. Zhu, T.~Park, P.~Isola, and A.~A. Efros.
\newblock Unpaired image-to-image translation using cycle-consistent
  adversarial networks.
\newblock {\em arXiv preprint arXiv:1703.10593}, 2017.

\end{thebibliography}
}

%
%
%
%

\end{document}